\documentclass{article}

\usepackage{microtype}
\usepackage{graphicx}
\usepackage{subfigure}
\usepackage{booktabs} 
\usepackage{multirow}
\usepackage{hyperref}

\usepackage[accepted]{icml2025}

\usepackage{amsmath}
\usepackage{amssymb}
\usepackage{mathtools}
\usepackage{amsthm}
\usepackage{algorithm}

\usepackage[capitalize,noabbrev]{cleveref}

\theoremstyle{plain}

\theoremstyle{definition}

\theoremstyle{remark}

\usepackage[textsize=tiny]{todonotes}

\icmltitlerunning{Submission and Formatting Instructions for ICML 2025}

\begin{document}

\twocolumn[
\icmltitle{Planning of Heuristics: Strategic Planning on Large Language Models with Monte Carlo Tree Search for Automating Heuristic Optimization}
\begin{icmlauthorlist}

\icmlauthor{Hui Wang}{st1}
\icmlauthor{Xufeng Zhang}{st1}
\icmlauthor{Chaoxu Mu}{st1,st2}


\end{icmlauthorlist}
\icmlaffiliation{st1}{School of Artificial Intelligence, Anhui University, Hefei, Anhui, China}
\icmlaffiliation{st2}{Pengcheng Laboratory, Shenzhen, Guangdong, China}

\icmlcorrespondingauthor{Chaoxu Mu}{cxmu@tju.edu.cn}



\icmlsetsymbol{equal}{*}

\icmlkeywords{Machine Learning, ICML}

\vskip 0.3in
]



\printAffiliationsAndNotice{Hui Wang and Xufeng Zhang contributed equally to this work. }
\begin{abstract}
Heuristics have achieved great success in solving combinatorial optimization problems~(COPs). However, heuristics designed by humans require too much domain knowledge and testing time. Since Large Language Models~(LLMs) possess strong capabilities to understand and generate content with a knowledge base that covers various domains, they offer potential ways to automatically optimize heuristics. To this end, we propose Planning of Heuristics~(PoH), an optimization method that integrates LLM self-reflection with Monte Carlo Tree Search, a well-known planning algorithm. PoH iteratively refines generated heuristics by evaluating their performance and providing improvement suggestions. Our method enables to iteratively evaluate the generated heuristics~(states) and improve them based on the improvement suggestions~(actions) and evaluation results~(rewards), by effectively simulating future states to search for paths with higher rewards. In this paper, we apply PoH to solve the Traveling Salesman Problem and the Flow Shop Scheduling Problem. The experimental results show that PoH outperforms hand-crafted heuristics and other Automatic Heuristic Design methods based on LLMs, and achieves the state-of-the-art performance in automating heuristic optimization with LLMs to solve tested COPs, especially with large sizes.
\end{abstract}

\section{Introduction}
Combinatorial optimization problems~(COPs) commonly exist in diverse fields such as national defense\cite{XING202457}, transportation\cite{wang2021deep}, industry\cite{zhao2024combinatorial}, and communication\cite{witt2024ilp}. Their substantial theoretical and practical value has made efficient solution of COPs a key research focus in both academia and industry. Heuristics are widely studied to solve COPs and achieve superior performance. The typical heuristics consist of guided local search~(GLS)~\cite{VOUDOURIS1999469}, genetic algorithm~(GA)~\cite{holland1973genetic}, and ant colony optimization~(ACO)~\cite{dorigo1996ant}. 
Usually, achieving satisfactory solutions with these methods requires human experts to manually adjust the heuristics for each specific problem. Although manually designed heuristics can be effective in many cases, this approach requires much time for experts to design, implement, and validate heuristics. In addition, for many complex COPs, this approach may result in errors. Consequently, Automatic Heuristic Design~(AHD) has emerged as a promising alternative~\cite{10.1145/3657604.3662042,10.5555/3586589.3586778}. The increasing availability of computational resources further facilitates AHD. By reducing the reliance on specialized domain knowledge, AHD enables us to explore large design spaces, demonstrating substantial potential to address complex COPs.

Given the fact that Large Language Models~(LLMs) have demonstrated remarkable capabilities in addressing COPs, due to their broad knowledge understanding and powerful reasoning abilities~\cite{yang2024large}. Furthermore, leveraging their extensive training corpora, LLMs benefit from a wider search space compared to traditional evolutionary computation~(EC) algorithms, leading to performance improvements~\cite{liu2024large,ma2024llamoco}. Recent research has successfully applied LLMs to heuristic generation and evolutionary search processes~\cite{Liu2024EvolutionOH,ye2024reevo,article}.

Therefore, in this paper, we introduce a novel AHD approach called Planning of Heuristics~(PoH). PoH reframes heuristic optimization as a strategic planning problem to manage the complexity of search spaces. Using strategic planning, PoH iteratively refines heuristics~(represented as states) through insightful improvement proposals~(actions). Starting with an initial heuristic~(state), the system systematically explores the heuristic search space using a tree-based search, prioritizing high-reward trajectories to efficiently navigate this extensive space. Using the Monte Carlo Tree Search~(MCTS) planning strategy, PoH can anticipate and simulate future rewards, subsequently using backpropagation to update reward values and guide the search toward more promising trajectories.

Above all, the main contributions of this paper can be summarized as follows. 
\begin{figure*}[t!]
    \centering
    \includegraphics[width=0.95\textwidth]{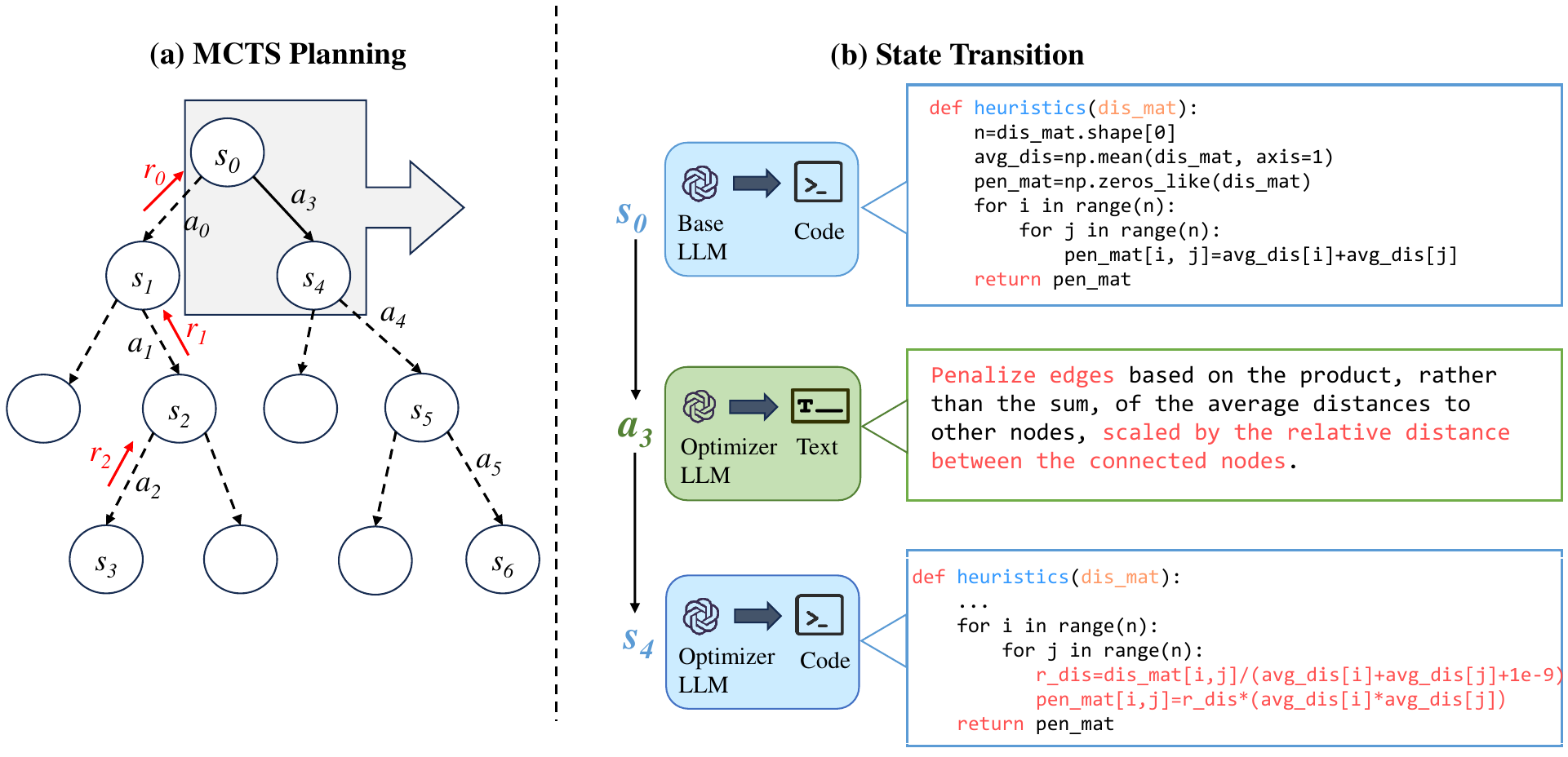}
    \caption{(a) MCTS planning for heuristic generation. The tree structure enables strategic planning of PoH . (b) A simplified state transition example. The base LLM first initialize~(generate) a current heuristic~(state), the optimizer LLM gathers improvement suggestions from the task dataset. An optimizer LLM then refines these suggestions, and updates the heuristic~(state) accordingly, transitioning to the next state.}
    \label{fig:framework}
\end{figure*}
\begin{itemize}
\item We propose PoH, a novel automated heuristic optimization method that leverages MCTS planning to strategically and efficiently explore the complex heuristic search space.

\item We show that PoH outperforms several existing automated heuristic design~(AHD) methods, including those utilizing LLMs for automatic heuristic optimization.

\item Unlike other frameworks that optimize heuristics with LLMs, PoH is a novel framework that combines iterative self-reflection with planning to optimize heuristics.
\end{itemize}

\section{Related Work}
\subsection{LLMs for Optimization}

LLMs have recently been used to address optimization problems through prompt engineering for specific issues~\cite{yang2024large,wei2022chain}. However, relying solely on prompt engineering has proven to have limited effectiveness in complex optimization scenarios. Inspired by the automatic generation of heuristics, researchers have explored combining evolutionary computation~(EC) with LLMs to generate and refine heuristics. FunSearch~\cite{article} presents a novel approach that searches within the function space, using LLMs to iteratively improve the quality of generated heuristics within an evolutionary framework. Evolution of Heuristics~(EoH)~\cite{Liu2024EvolutionOH} employs natural language to represent heuristic ideas; LLMs first generate natural language descriptions of heuristics, which are then used to produce executable heuristic code. This evolutionary search framework allows for the simultaneous improvement of both the descriptions and the code, contributing to EoH's effectiveness and efficiency. Similarly, Reflective Evolution (ReEvo)~\cite{ye2024reevo} enhances the efficiency of heuristic evolution by combining evolutionary search with the self-reflection capabilities of LLMs.
\subsection{LLMs with Self-reflection and Planning}

Self-reflection is a cognitive process where an individual contemplates their own thoughts, feelings, and actions, enabling the recognition of mistakes during problem-solving and the continuous adjustment of strategies. Similarly, guiding LLMs to engage in self-reflection, allowing them to evaluate their generated content, can effectively improve their problem solving performance~\cite{shinn2023reflexionlanguageagentsverbal,10.1145/3657604.3662042,huang2022innermonologueembodiedreasoning}.

Planning is a crucial tool for agents operating in complex and dynamic environments and making high-quality decisions. Traditional planning methods, which represent problems in a structured format, can leverage efficient search algorithms to generate the correct and optimal solutions~\cite{cheng-etal-2022-improving,vath-etal-2023-conversational,Liu2023LLMPEL}. This motivates research that combines LLMs with planning techniques, such as using MCTS to explore more comprehensive reasoning paths~\cite{wang2024promptagent,hao2023reasoning}.

However, existing LLM-based heuristic optimization methods often rely on evolutionary frameworks or incorporate reflection mechanisms, but typically employ direct iterative algorithms without principled strategies for guided exploration.  In contrast, our proposed PoH method synergistically combines self-reflection with planning, leveraging MCTS to efficiently search the vast heuristic space.

\section{Methodology}
This section introduces PoH, a framework that empowers LLMs to strategically plan a coherent reasoning trajectory to solve a wide range of COPs. We first format the definition of the heuristic optimization problem and then present the proposed PoH framework. Finally, we describe how MCTS planning is employed to effectively explore the vast heuristic space and to identify optimal trajectories.

\subsection{Problem Definition}
Following a standard setting in heuristic optimization, COP is defined by a solution space $\mathcal{S}$ and an objective function $f : \mathcal S \mapsto \mathcal R$. Heuristic optimization typically searches within a heuristic space ${H}$ to find an optimal heuristic ${H}^*$ that minimizes an evaluation function, formally expressed as ${H}^*=\arg\min_{{h}\in{H}}F(h)$. Unlike traditional methods of optimizing heuristics, we leverage LLMs to generate heuristics, enabling exploration of this open heuristic space. Specifically, we evaluate the generated heuristic $H$ on a training set $I$ to maximize performance according to a reward function $\mathcal{R}$, which can be formulated as ${H}^*=\arg\max_{{h}\in{H}}\mathcal{R}(p_{\mathcal{B}}({I},{H}))$.

\subsection{PoH Framework}
\begin{algorithm}[!t] 
\caption{{Planning of Heuristics Method}}
\label{alg:algorithm}
\textbf{Inputs}:Initial state $s_0$, state transition function $p_\theta(s'|s, a)$, reward function $r_\theta(s, a)$, action generator $p_\alpha(a|s)$, number of generated actions $h$, depth limit $l$, number of iterations $I$, exploration weight $e$\\
\textbf{Initialize}: memory of actions $A : \mathcal S \mapsto \mathcal A$, children $\text{c} : \mathcal S \times \mathcal A \mapsto \mathcal S$, rewards $r : \mathcal S \times \mathcal A \mapsto \mathbb R$, State-Action value function $Q : \mathcal S \times \mathcal A \mapsto \mathbb R$, visited count $N : \mathcal S \mapsto \mathbb N$
\begin{algorithmic}[1]  
    \FOR {$n \gets 0, \dots, I - 1$}
        \FOR {$t \gets 0, \dots, l - 1$}
            \IF{$A(s_t)$ is not empty} 
            \STATE $a_t\gets\arg\max_{a\in{A}(s_t)} \left[ Q(s_t, a) + e\cdot \sqrt{\frac{\ln N(s_t)}{N(\text{c}(s_t, a))}} \right]$ 
            \STATE $s_{t + 1} \gets \text{c}(s_t, a_t)$, $r_t \gets r(s_t, a_t)$,
            \STATE $N(s_{t}) \gets N(s_{t}) + 1$ 
            \ELSE 
                \FOR {$j \gets 1, \dots, h$}
                \STATE Sample $a_t^j \sim p_\alpha(a | s_t)$, $s_{t+1}^j \sim p_\theta(s | s_t, a_t^j)$,
                \STATE $r_t^i \gets r_\theta(s_t, a_t^j)$
                \STATE Update $A(s_t) \gets \{a_t^j\}_{j=1}^d$,
                \STATE $\text{c}(s_t, a_t^j) \gets s_{t+1}^j$,
                \STATE $r(s_t, a_t^j) \gets r_t^j$
                \ENDFOR
            \STATE $a_t \gets \arg\max_{a^i_t \in {A}(s_t)} r_t^j (s_t, a^j_t) $ 
            \STATE $s_{t + 1} \gets \text{c}(s_t, a_t)$, $r_t \gets r(s_t, a_t)$,
            \STATE $N(s_{t}) \gets N(s_{t}) + 1$ 
            \ENDIF
            \STATE \algorithmicif\ {$s_{t+1}$ is an early-stopping state}\ \algorithmicthen\ \textbf{break}
        \ENDFOR
        \STATE $T' \gets$ the actual number of steps
        \FOR {$t \gets T' - 1, \dots, 0$} 
            \STATE Update $Q(s_t, a_t)$ with $\{r_t, r_{t+1}, \dots, r_l\}$ 
        \ENDFOR
    \ENDFOR
\end{algorithmic}
\end{algorithm}
PoH regards the heuristic optimization problem as a Markov Decision Process~(MDP), defined by the tuple ~$(\mathcal{S}, \mathcal{A}, \mathcal{T},\mathcal{R})$, with a defined state and action space. Here, $\mathcal{S}$ represents the state space, $\mathcal{A}$ is the action space, $\mathcal{T}$ denotes the transition function, $\mathcal{T}: \mathcal{S}\times \mathcal{A}\mapsto \mathcal{S}$, and $\mathcal{R}$ is the reward function $\mathcal{R}: \mathcal{S} \times \mathcal{A} \mapsto \mathbb{R}$. In our work, we formulate the MDP concretely as follows.

\textbf{State} $\mathcal{S}$ is a set of generated heuristics, where $s_t\in \mathcal{S}$, and $s_t$ is the generated heuristic at step $t$.

\textbf{Action} $\mathcal{A}$ is a set of improvement suggestions, where $a_t\in \mathcal{A}$ and $a_t$ is the generated improvement suggestion for heuristic $s_t$.

\textbf{Transition function} $\mathcal{T}$: When applying an action $a_t$ to state $s_t$, it yields $s_{t+1}$, which means the heuristic~(state $s_t$) accepts the improvement suggestion~(action $a_t$) and updates to a new heuristic~(state $s_{t+1}$).

\textbf{Reward} $\mathcal{R}$ is evaluated as $1 - \frac{O_{st} - O_{best}}{O_{best}}$ , where $O_{best}$ is known instance-specific baselines, $O_{st}$ is the current heuristic.

As illustrated in Figure~\ref{fig:framework} (a), given a current state $s_t$,  PoH iteratively generates an action $a_t$ according to $a_t \sim p_{\theta}(a | s_t)$, $p_{{B}}(a | s_t)$ represents that the base LLM $p_{\theta}$ generates improvement actions $a$ given the current state $s_t$. The action generation process, detailed in Figure~\ref{fig:framework} (b), comprises two steps: first, evaluating the heuristic~(state) generated by the base model. This evaluation involves replacing the distance matrix update function of the GLS search algorithm with the generated heuristic and then evaluating it on the training set to obtain the current heuristic's optimal gap~(\%). Subsequently, for action generation, the optimizer LLM $p_{\mathcal{B}}$ generates improvement suggestions $a_t$ based on the current heuristic $s_t$ and the optimal gap~(\%). PoH then determines the subsequent state using the transition function $p_{\mathcal{\theta}}(s_{t+1} | s_t, a_t)$ to update the heuristic. Given the current improvement suggestions~(action), the optimizer LLM $p_{\mathcal{\theta}}$ generates a new heuristic~(state) that incorporates relevant domain knowledge and effectively addresses model suggestions, which is similar to how AHD via Hyper-Heuristics updates algorithms based on improvement suggestions.

The quality of applying the action $a_t$ to state $s_t$ is then evaluated by the reward function $r_t = r(s_t, a_t)$. The reward function is defined as 1 minus the percentage gap between the solution and the optimal solution for an instance, aiming to maximize solution quality and minimize the gap with the optimal solution.

\subsection{Planning with Monte Carlo Tree Search}
PoH incorporates MCTS as its strategic planning method. A typical MCTS loop consists of the following 4 steps. 

\noindent \textbf{Selection} 
Starting from the root node in a search tree, the selection phase traverses the child nodes according to the Upper Confidence Bound~(UCT) formula, which balances exploitation and exploration, until a leaf node is reached. The UCT formula is as follows:
\vspace{-5pt}
{
\small
\begin{align}
    &a^*=\arg\max_{a\in A(s)}\left[Q(s,a)+e\sqrt{\frac{\ln N(s)}{N(c(s,a))}}\right]&  
    \label{eq:1}
\end{align}
}
Here, ${A}(s)$ represents the set of actions available at node $s$, $N(s)$ represents the number of times node $s$ has been visited, $\text{c}(s, a)$ represents the child node resulting from applying action $a$ to node $s$, and $e$ is a constant that adjusts the degree of exploration. The first term of the formula, $Q(s,a)$, reflects exploitation, while the second term reflects exploration, quantifying the uncertainty associated with the nodes visited less frequently. Specifically, if a node and its child node have been explored insufficiently, the value of the second term will be higher.

\noindent \textbf{Expansion} When a leaf node is reached and a terminal state has not yet been achieved, the expansion phase creates one or more new nodes. These new nodes represent future states that may be reached from the current state. During the expansion phase, no evaluation or simulation of these newly added child nodes takes place. Expansion merely increases the structure of the tree in preparation for subsequent simulation steps.

\noindent \textbf{Simulation} Starting from the current node, the simulation phase selects actions from all possible actions according to a policy, resulting in state transitions. If a terminal state is not reached, actions continue to be selected until a terminal state is reached. 

\noindent \textbf{Back-propagation} Back-propagation propagates the simulation result back through the search tree to update node information. When the simulation reaches a terminal state, the $Q$ value of the current node is updated. By continuously updating node information, MCTS guides the search process in more promising directions.

\vspace{-5pt}
{
\small
\begin{align}
    Q (s_t, a_t) = \max_{s_t, a_t, r_t, \dots, s_l, a_l, r_l, s_{l+1}} \operatorname{avg}(r_t, \dots, r_l). 
    \label{eq:2}
\end{align}
}

Above all, the pseudo code of the PoH method is given as Algorithm~\ref{alg:algorithm}. The terminal state is when it reaches the predefined maximum depth or meets the early-stopping
condition~(Line 18 in Algorithm~\ref{alg:algorithm}). The early stopping condition is activated when the state's reward is below a minimum threshold
or above a maximum threshold. Specifically, the minimum threshold is the average of the rewards of the parent node and the
root node, while the maximum threshold is the maximum value among all current nodes. PoH iteratively executes these four phases to explore the heuristic space. The algorithm terminates when a predefined number of iterations is reached, and the best trajectory and corresponding node are then selected for evaluation.

\section{Experiments}

\subsection{Experiments Settings}    
\noindent \textbf{Tasks and Datasets} We evaluated PoH on two classical combinatorial optimization problems: The TSP and the FSSP. The TSP seeks the shortest route that starts at a depot, visits all customer locations exactly once, and returns to the depot. As a canonical combinatorial optimization problem, it serves as a standard benchmark for heuristic methods. Following the setup in~\cite{Liu2024EvolutionOH}, we performed heuristic optimization on a training set of 64 TSP instances, each with 100 city nodes randomly distributed in $[0,1]^2$~\cite{kool2019attentionlearnsolverouting}. We used the average optimal gap, calculated with respect to the solutions generated by Concorde~\cite{applegate2006concorde}, as the evaluation function.

 FSSP is a classical problem in production scheduling, which involves scheduling of $n$ jobs on $m$ machines. Each job consists of $m$ operations that must be processed on the machines in the same predetermined order. Each job may have different processing times on each machine. A job can only begin processing on the next machine after completing processing on the previous machine. Each machine can process only one job at a time, and each job can be processed on only one machine at a time. The objective is to minimize the makespan, the total time of finishing all jobs. For heuristic optimization, we used a training set of 64 randomly generated FSSP instances, each with 50 jobs and a varying number of machines~(from 2 to 20), consistent with~\cite{Liu2024EvolutionOH}. The processing times of the jobs were randomly generated from a uniform distribution between 0 and 1~\cite{pan2021deep}. The average optimal makespan, calculated with respect to the solutions generated by~\cite{Liu2024EvolutionOH}, served as the evaluation function.

\noindent\textbf{Baselines} We compare PoH with three categories of baselines: Guided Local Search~(GLS) algorithms, LLM-generated heuristics, and other algorithms for the TSP. GLS algorithms include Knowledge-Guided Local Search (KGLS)~\cite{arnold2019_KGLS_VRP}, GNNGLS~\cite{hudson2022_gnn_gls}, NeuralGLS~\cite{sui2023neuralgls}, and a state-of-the-art neural combinatorial optimization~(NCO) method~\cite{luo2023neural}. LLM-generated heuristics include EoH~\cite{Liu2024EvolutionOH} and ReEvo\cite{ye2024reevo}. Other algorithms include Google Or-Tools~\cite{ortools}, the Attention Model (AM)~\cite{kool2019attentionlearnsolverouting}, POMO~\cite{kwon2021matrix}, and LEHD~\cite{luo2023neural}.

For FSSP, we compared PoH with NEH~\cite{nawaz1983heuristic}, NEHFF~\cite{fernandez2014insertion}, Local Search~(LS), Iterated Local Search~(ILS)~\cite{Sttzle1998ApplyingIL}, and the AHD methods PFSPNet and PFSPNet\_NEH~\cite{pan2021deep}. NEH and NEHFF are widely recognized as efficient heuristics for this problem. LS and ILS are classical search methods for the FSSP; we used the same operators in LS as those used in PoH.

\noindent\textbf{Implementation details}
We applied PoH to automatically design Guided Local Search~(GLS) heuristics for both the TSP and the FSSP. GLS aims to guide local search away from local optima. A key challenge is to update the objective function to direct the search toward more promising regions. Because the search space is primarily defined by the distance matrix~(for the TSP) or the time matrix~(for the FSSP), PoH aims to discover heuristics that efficiently update these matrices.

Specifically, for the TSP, following~\cite{arnold2019knowledge}, the input to the generated heuristic is a distance matrix, and the output is an updated distance matrix. GLS combines a perturbation phase~(where edges with higher heuristic values are preferentially penalized) with a local search operator applied to the updated search space.

For the FSSP, we adopted the same GLS framework used for the TSP, employing PoH to generate a heuristic that updates the execution time matrix and determines the jobs to be perturbed. We used the \textbf{Swap} and \textbf{Relocate} local search operators. The inputs to this heuristic are the number of jobs and machines, the current job sequence, and the time matrix; the outputs are the updated job perturbation priority and the updated time matrix.

In our experiments, we used GPT-4.0 as both the base and optimizer models, with temperatures of 0.0 and 1.0, respectively. The MCTS parameters were set as follows: 10 iterations, an expansion width of 5, a maximum depth of 5, and an exploration weight of 2.5.

\subsection{Result and Analysis}

\noindent\textbf{Traveling Salesman Problem} We firstly evaluated the performance of PoH and several other well-studied methods on common instances from TSPLIB~\cite{reinelt1991tsplib}. Table~\ref{tab:tsplib} presents the relative distance (\%) to the best-known solutions for these methods in a typical subset~(selected by~\cite{Liu2024EvolutionOH}) of instances. The complete comparison results with additional TSPLIB instances and other algorithms are provided in the Appendix. As shown in Table~\ref{tab:tsplib}, EoH achieved the best results on several instances, finding the optimal solution for pr124, kroA150 and u159. In addition, PoH consistently outperformed all other algorithms on all tested instances, demonstrating exceptional robustness and effectiveness, particularly on the larger instance kroB200, where it consistently produced near-optimal solutions.

\begin{table}[!t]
\centering
\small
\caption{\textbf{Traveling salesman problem results.} Comparison of the relative distance (\%) to the best-known solutions~(lower is better) for various routing heuristics on a subset of TSPLib instances. \label{tab:tsplib}}
\renewcommand{\arraystretch}{1.25}
\resizebox{.49\textwidth}{!}{%
\begin{tabular}{@{\hspace{1mm}}lcccccc@{\hspace{1mm}}}
\toprule
Method & rd100 & pr124 & bier127 & kroA150 & u159  & kroB200 \\
\midrule
Or-Tools & 0.01  & 0.55  & 0.66    & 0.02    & 1.75  & 2.57    \\
\midrule
AM       & 3.41  & 3.68  & 5.91    & 3.78    & 7.55  & 7.11    \\
POMO     & 0.01  & 0.60  & 13.72   & 0.70    & 0.95  & 1.58    \\
LEHD     & 0.01  & 1.11  & 4.76    & 1.40    & 1.13  & 0.64    \\
\midrule
GNNGLS & 0.46  & 0.76  & 1.95    & 2.98    & 1.02  & 1.53    \\
GLS & 0.01  & 0.60  & 0.59    & 1.75    & 0.74  & 1.43    \\
KGLS & 0.01  & 0.08  & 0.42    & 0.17    & 0.96  & 0.89    \\
\midrule
EoH     & \textbf{0.01}  & \textbf{0.00}  & 0.42    & \textbf{0.00 }   & \textbf{0.00 } & 0.20  \\
\textbf{PoH}~(ours)     & \textbf{0.01}  & \textbf{0.00}  & \textbf{0.01}    & \textbf{0.00 }   & \textbf{0.00 } & \textbf{0.01}   \\
\bottomrule
\end{tabular}%
}
\end{table}

\begin{figure}[!t]
		\centering
		\includegraphics[width=1.0\linewidth]{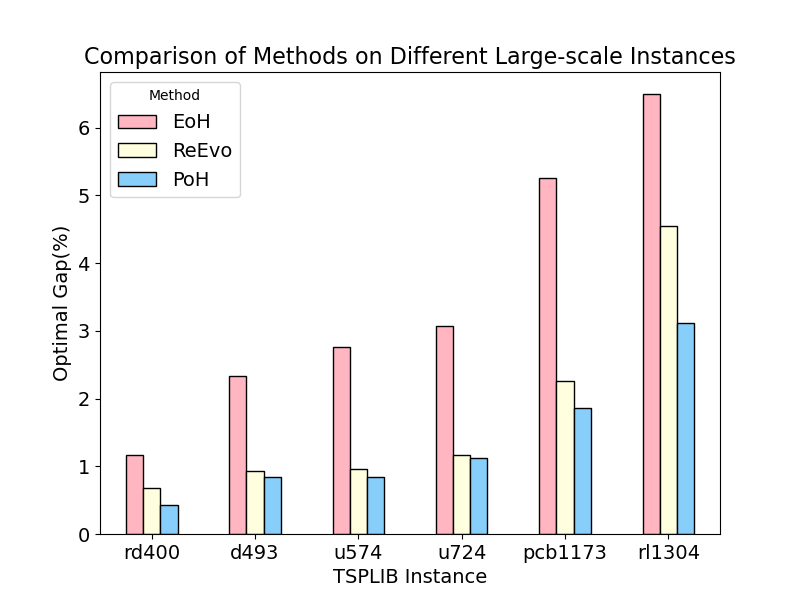}
		\caption{Comparative analysis of LLM-enhanced heuristics on large-scale TSPLIB instances by showing the optimal gap relative to the optimal solution. PoH achieved significantly smallest gaps especially on problems with larger sizes, which indicates the promising potential of PoH to solve large-scale COPs.}
		\label{fig:large_scale}
	\end{figure}

Figure~\ref{fig:large_scale} compares PoH with two other LLM-based heuristic optimization methods on several large-scale TSPLIB instances. The results demonstrate that PoH significantly outperforms the other two methods on all tested instances. Moreover, PoH's advantage becomes more pronounced as the problem size increases, indicating its superior ability to find near-optimal solutions for larger instances. Although ReEvo exhibits a relatively small difference compared to PoH on instances with fewer than 1000 nodes, the advantage of PoH is obvious on larger instances.

\begin{table*}[!t]
    \caption{Evaluation results of different local search (LS) variants. We report optimality gaps and per-instance execution time.}
    \renewcommand{\arraystretch}{1.25}
    \label{tab: main-exp-gls}
    \resizebox{\textwidth}{!}{
    \begin{tabular}{c|c|cc|cc|cc|cc}
    \toprule
    \multirow{2}{*}{Method} & \multirow{2}{*}{Type} & \multicolumn{2}{c|}{TSP20} & \multicolumn{2}{c|}{TSP50} & \multicolumn{2}{c|}{TSP100} & \multicolumn{2}{c}{TSP200} \\
                            &                       & Opt. gap (\%)  & Time (s) & Opt. gap (\%)  & Time (s) & Opt. gap (\%)  & Time (s)  & Opt. gap (\%)  & Time (s)  \\
    \midrule
    NeuOpt & LS+RL & 0.000  & 0.124  & 0.000 & 1.32 & 0.027 & 2.67  & 0.403  & 4.81      \\
    GNNGLS  & GLS+SL  & 0.000 & 0.116  & 0.052 & 3.83  & 0.705 & 6.78  & 3.522   & 9.92      \\
    NeuralGLS  & GLS+SL  & 0.000 & 10.005 & 0.003 & 10.01 & 0.470 & 10.02 & 3.622 & 10.12 \\
    KGLS    & GLS     &  0.004     &    0.001    &   0.017    &   0.03     &  0.002     &   1.55     &    0.284   &     2.52   \\
   \midrule
    EoH      & GLS+LLMs & 0.000 & 0.563    & 0.000 & 1.90    & 0.025 & 5.87    & 0.338     & 17.52      \\ 
    ReEvo & GLS+LLMs &   0.000    &     0.001   &   0.000    &    0.02    &   0.000   &   0.95     &  0.306    &   1.70     \\
    \textbf{PoH}~(ours) & GLS+LLMs &   \textbf{0.000} &\textbf{0.001}     &   \textbf{0.000}  &\textbf{0.02}     &   \textbf{0.000}  &\textbf{0.95}     &  \textbf{0.227}   &\textbf{1.58}     \\
    \bottomrule
    \end{tabular}}
    \end{table*}
    
In addition, we also conducted experiments on the 64-instance dataset generated by ReEvo~\cite{ye2024reevo} using seed 1234. The results, presented in Table~\ref{tab: main-exp-gls}, demonstrate the performance achieved by PoH across different sizes of TSP instances. PoH achieves optimal solutions~(0.000\% optimality gap) on TSP20, TSP50, and TSP100 instances, matching the best-performing baselines on these smaller instances. More importantly, on larger TSP200 instances, PoH achieves a significantly lower optimality gap of 0.227\% compared to all other methods, including both GLS-based methods and other LLM-generated heuristics. This result highlights PoH's superior ability to scale to larger and more complex TSP instances. Furthermore, PoH achieves these impressive results with competitive execution times, as shown in Table~\ref{tab: main-exp-gls}. Notably, on TSP200, PoH achieves its superior optimality gap with a runtime comparable to ReEvo, and a runtime much shorter than EoH.

\noindent\textbf{Flow shop scheduling problem}
As shown in Table~\ref{tab:fssp}, PoH demonstrates strong overall performance in all Taillard instances, achieving the lowest average relative makespan in four of the six combinations tested (n20m10, n50m10, n100m10 and n100m20). In particular, PoH achieves significant improvements over traditional methods such as NEH, NEHFF, LS, and ILS1, as well as the deep learning-based methods PFSPNet and PFSPNet\_NEH, particularly on instances with 100 jobs. For example, on n100m10, PoH achieves a makespan of 0.11\%, significantly lower than the next best result of 0.14\% achieved by EoH. Although EoH performs competitively on some smaller instances (n20m20 and n50m10), PoH matches or outperforms it in these cases. On the n50m20 instance, PoH achieves a makespan of 0.49\%, slightly higher than the 0.19\% achieved by EoH and the 0.47\% achieved by LS. However, PoH's consistent strong performance across the other instances, especially the larger ones, highlights its robustness. The substantial difference in performance between PoH and the deep learning methods PFSPNet and PFSPNet\_NEH further underscores the effectiveness of the PoH approach. More detailed results and analysis are provided in the Appendix.

\begin{table}[ht]
\centering
\large
\renewcommand{\arraystretch}{1.35}
\caption{\textbf{Flow shop scheduling problem results.} The comparison of the average relative makespan (\%) of various baselines on FSSP, evaluated on Taillard instances~(lower values indicate better performance).}
\resizebox{.48\textwidth}{!}{%
\begin{tabular}{ccccccc}
\toprule
Method & n20m10 & n20m20 & n50m10 & n50m20 & n100m10 & n100m20 \\
 \midrule
NEH          & 4.05   & 3.06   & 3.47   & 5.48   & 2.07    & 3.58  \\ 
NEHFF        & 4.15   & 2.72   & 3.62   & 5.10   & 1.88    & 3.73  \\
\midrule
PFSPNet      & 14.78  & 14.69  & 11.95  & 16.95  & 8.21    & 16.47  \\ 
PFSPNet\_NEH & 4.04   & 2.96   & 3.48   & 5.05   & 1.72    & 3.56  \\ 
\midrule
LS      & 2.77  & 2.60  & 3.33  & 4.67  & 1.38    & 3.51  \\ 
ILS1 & 0.33   & 0.29   & 1.47   & 2.13   & 0.77    & 2.27  \\ 
\midrule
EoH   & 0.30   & \textbf{0.10}   & \textbf{0.19}   & 0.60   &0.14    & 0.41   \\
\textbf{PoH}~(ours)   & \textbf{0.19}   & 0.22   & \textbf{0.19}   & \textbf{0.49}   & \textbf{0.11}    & \textbf{0.38}   \\
\bottomrule
\label{tab:fssp}
\end{tabular}%
}
\end{table} 

\subsection{Heuristic Generalization}
We compared PoH's performance using four commonly used LLMs: GPT-4.0, Gemini-1.5-Pro, GLM-4-Plus, and GPT-3.5-turbo to further demonstrate the generalization of our method across different LLMs. We conducted three independent runs for each LLM on the TSP200 test set, using the same experimental setup for all runs. The results are presented in Table~\ref{tab:LLMs comparision}. Among the LLMs tested, GPT-4.0 achieved the best performance with PoH, achieving a minimum gap of 0.227\% relative to the optimal solution in a single run and an average gap of 0.233\% across 3 runs.

\begin{table}[t]
\centering
\caption{Comparison of PoH with different LLMs. The average gap (\%) to the optimal solution on TSP200 instances.}
\renewcommand{\arraystretch}{1.2}
\resizebox{0.48\textwidth}{!}{%
\begin{tabular}{llllll}
\toprule
Method & LLM     & Run 1 & Run 2 & Run 3 & Average \\
\midrule
Sampling      & GPT-4.0& 0.762 & 0.790 & 0.822  & 0.791  \\
PoH    & GPT-3.5-turbo & 0.252  & 0.246  & 0.234 & 0.244        \\
PoH    & GLM-4-Plus      & 0.276& 0.254  & 0.260  &0.260      \\
PoH    & Gemini-1.5-Pro    & 0.233  & 0.236  &0.239  & 0.236         \\
PoH    & GPT-4.0        & 0.235  & \textbf{0.227}  & 0.238  & \textbf{0.233}\\
\bottomrule
\end{tabular}%
}
\label{tab:LLMs comparision}
\end{table}

\subsection{Ablation on Search Strategies}
\begin{figure}[!t]
		\centering
		\includegraphics[width=0.9\linewidth]{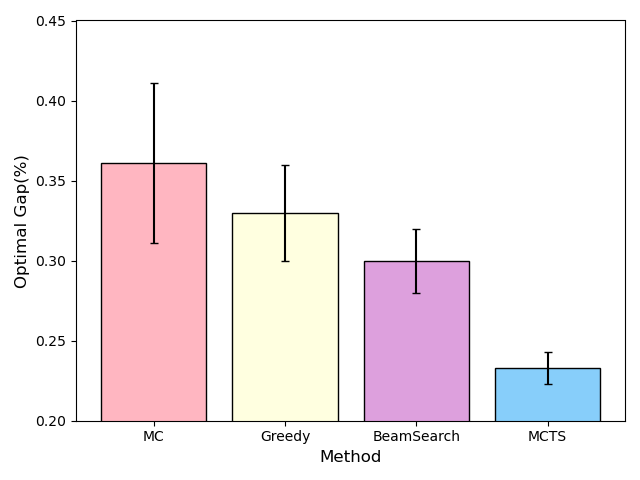}
		\caption{Ablation study on search methods on TSP200. We compare the optimal gap of four different search methods on the TSP200 test set. The horizontal axis represents the different methods, and the vertical axis represents the percentage of the optimal gap. The error bars indicate the standard deviation of the data, reflecting the stability of the results.}
		\label{fig:ablation}
	\end{figure}
We conducted ablation experiments to further investigate the impact of different search algorithms within the PoH framework. Specifically, we compared PoH using Monte Carlo~(MC) search, depth-first Greedy Search~(Greedy), and beam search. For each search algorithm, we maintain identical state transitions and action generation processes, only substituting MCTS with the respective algorithm. MC randomly samples and selects an action. Greedy selects the action with the highest immediate reward at each step. The beam search retains the most promising paths (determined by the beam width) during action selection and expands along these paths to find a solution. The total number of explored heuristics was kept constant in all search algorithms.

Figure~\ref{fig:ablation} demonstrates that both Greedy and Beam search outperform MC within the PoH framework, indicating PoH's capacity for iterative improvement through guided exploration. Furthermore, the beam search performs better than Greedy. In particular, MCTS achieves the smallest deviation from the optimal solution and exhibits greater stability~(smaller error bars).

 \begin{figure}[!b]
		\centering
		\includegraphics[width=\linewidth]{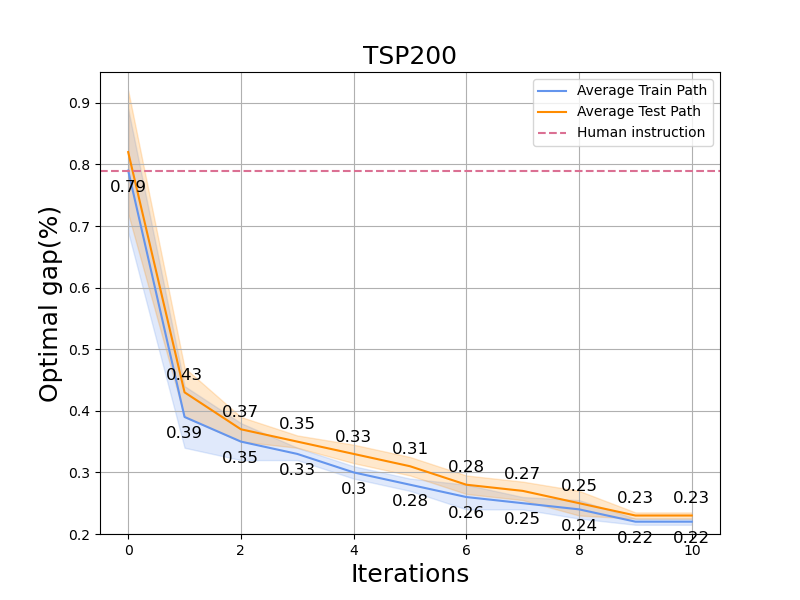}
		\caption{Convergence curves on TSP200 with varying iterations. The vertical axis represents the optimal gap to the baseline for TSP200. The horizontal axis represents the number of MCTS iterations. The shaded areas represent the standard deviation of the current node's performance.}
		\label{fig:iterations}
	\end{figure}
    
\begin{table}[!t]
\centering
\caption{Comparison of exploration efficiency with different search strategy. The average gap (\%) to the optimal solution on TSP200 instances. \label{tab:LLM}}
\renewcommand{\arraystretch}{1.2}
\resizebox{0.48\textwidth}{!}{%
\begin{tabular}{lcc}
\toprule
Method & Explored heuristics     & gap(\%)  \\
\midrule
Greedy Search      &34 & 0.530  \\
Beam Search    & 72 & 0.260          \\
MCTS~(our PoH)    & 60        &\textbf{ 0.227 }     \\
\bottomrule
\end{tabular}%
}
\end{table}

	\begin{figure}[!b]
		\centering
		\includegraphics[width=\linewidth]{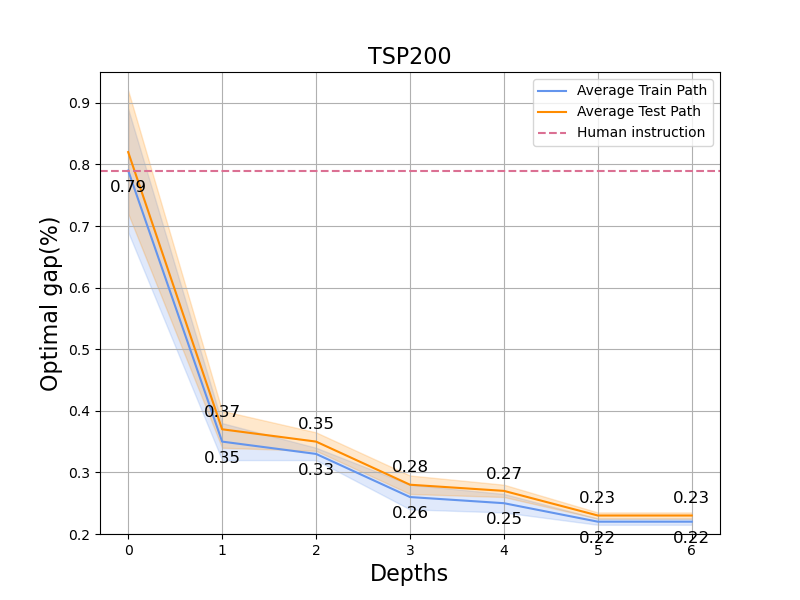}
		\caption{Convergence curves on TSP200 with varying tree depths. The vertical axis represents the optimal gap to the baseline for TSP200. The horizontal axis represents the MCTS tree depth. The shaded areas represent the standard deviation of the current node's performance.}
		\label{fig:depths}
	\end{figure}
    
\subsection{Exploration Efficiency Analysis}

To demonstrate PoH's effectiveness in exploring the prompt space through strategic planning, we compared it with several other search algorithms to analyze exploration efficiency. Specifically, we compared PoH with Greedy Search~(equivalent to Greedy in ablation study but with a expand width of 1) and beam search~(with a beam width of 3). These two search algorithms explored 34 and 72 heuristics, respectively. As shown in Table~\ref{tab:LLM}, increasing the beam width from 1 to 3 improves performance, but neither surpasses MCTS~(used in our proposed PoH). Furthermore, beam search requires a higher computational cost. These results demonstrate that PoH strategically and efficiently traverses the heuristic space and effectively finds optimal trajectories.

\subsection{Convergence Analysis}
   \begin{figure*}[!ht]
    \centering
    \includegraphics[width=0.9\textwidth]{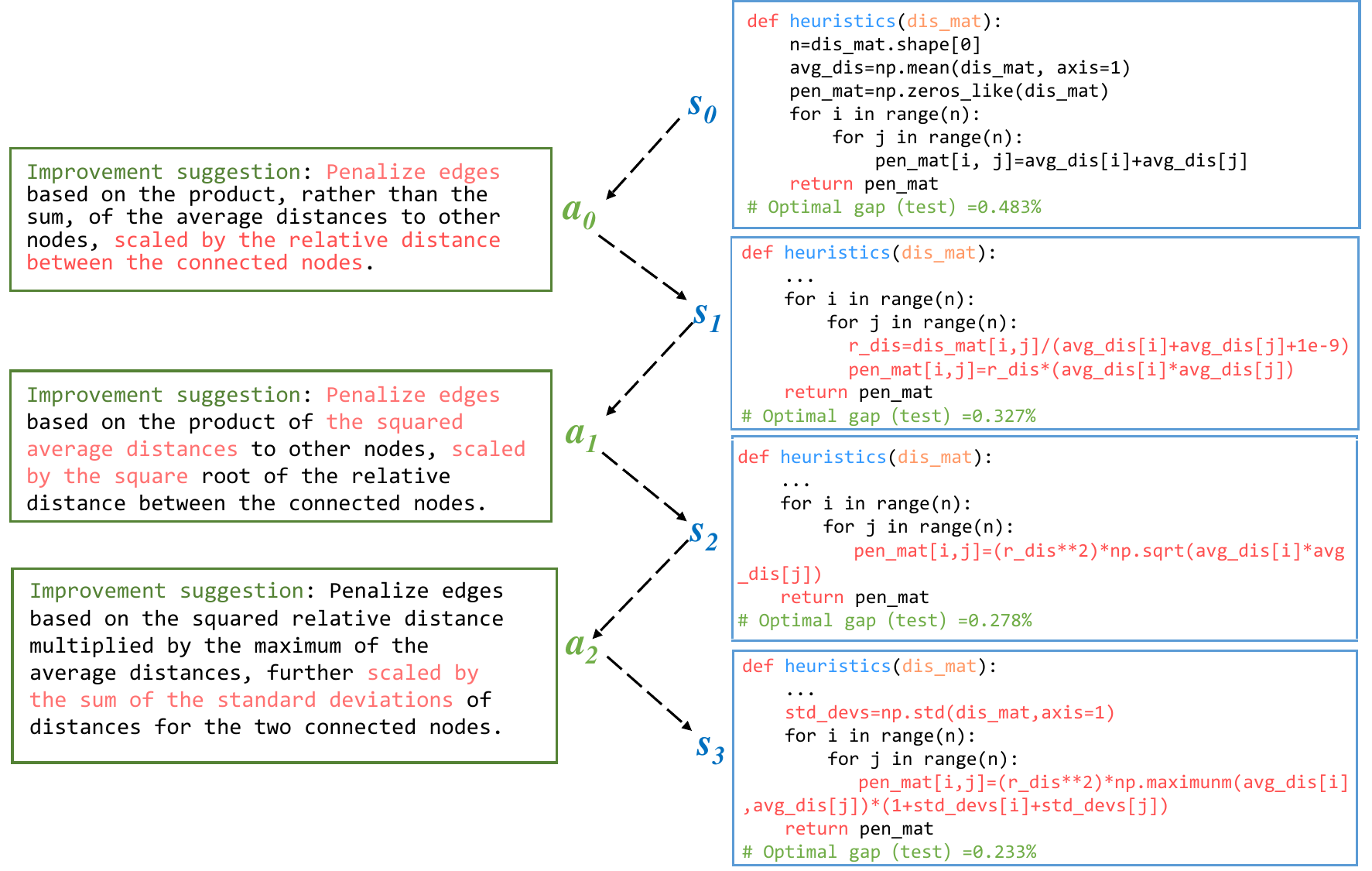}
    \caption{This figure illustrates the state-action transition process of MCTS within PoH for training on the TSP. An initial heuristic~(state), $s_0$, is generated by a base model. In each transition, an optimizer model proposes improvement suggestions~(actions) based on the current state. PoH then generates a new heuristic~(state) considering both the previous state and these suggestions. Evaluated on TSP200 test instances, this process reduced the gap between the obtained solution and the optimal solution from 0.483\% to 0.233\%.}
    \label{fig:PoH framework}
\end{figure*}

We conducted a convergence analysis of PoH's learning process, visualizing the optimal gap with respect to both the number of MCTS iterations and tree depth. We used TSP200 as a representative test instance for clearer visualization. Figure~\ref{fig:iterations} shows the convergence with varying iterations, where the horizontal axis represents the number of iterations and the vertical axis represents the optimal gap. The performance of PoH during training and testing is shown by blue and yellow lines, respectively. Figure~\ref{fig:depths} shows the convergence with varying tree depths, where the horizontal axis represents the tree depth and the vertical axis represents the optimal gap. In both figures, the optimal gap decreases as the number of iterations and tree depth increase, demonstrating PoH's ability to iteratively improve the generated heuristic.

\subsection{Qualitative Analysis}
Figure~\ref{fig:PoH framework} provides a qualitative analysis of how PoH iteratively refines its heuristic~(state) based on improvement suggestions~(actions), demonstrating its capacity for strategic planning. The figure depicts four states~($s_0$ to $s_3$) and three actions~($a_0$ to $a_2$) within a TSP training trajectory, showcasing the continuous optimization of the heuristic from its initial state~($s_0$). Each subsequent state incorporates the suggestions from previous iterations, resulting in a heuristic with a progressively decreasing gap from the optimal solution on the test instance, reducing from 0.483\% to 0.233\%.

\section{Conclusion}
This paper proposed Planning of Heuristics~(PoH), a novel framework for automated heuristic design that combines Large Language Models~(LLMs) with the planning strategies of MCTS. PoH strategically searches within the vast heuristic space through MCTS-based planning algorithm. Furthermore, PoH leverages LLM self-reflection to identify shortcomings in generated heuristics and propose targeted improvements. We evaluated PoH on two benchmark combinatorial optimization problems: the Traveling Salesman Problem~(TSP) and the Flow Shop Scheduling Problem~(FSSP). Experimental results demonstrate that PoH outperforms both other LLM-optimized heuristics and manually designed heuristics.

For future work, researchers are encouraged to apply our proposed method to solve large-scale COPs in practice. MCTS simulation may cost much time, incorporating cheaper simulation methods such as Gumbel MCTS should be promising on sovling large-scale COPs .

\section{Acknowledgment}
This paper presents work whose goal is to advance the field of Machine Learning. There are many potential societal consequences of our work, none of which we feel must be specifically highlighted here.
\bibliography{example_paper}
\bibliographystyle{icml2025}

\newpage
\onecolumn
\section{Appendix}
\subsection{Implementation details}
\textbf{Planning of Heuristic~(PoH)}. PoH performs MCTS planning within the space of heuristics. MCTS is a search algorithm designed for complex decision-making problems, particularly those with vast and difficult-to-enumerate state spaces. Its core principle involves exploring potential decision paths through simulated random strategies and evaluating the potential of each node using statistical methods. In PoH, the terminal state conditions and the reward function are key components. A terminal state is reached when the length of the explored path reaches a predefined depth limit. The reward function is derived from the error achieved by the heuristic generated when evaluated on a validation dataset.

PoH leverages large language models (LLMs) to generate an initial heuristic based on prompts, using this heuristic as the initial state and the root node for subsequent expansion. The agent performs 10 MCTS iterations, each consisting of four key phases: selection, expansion, simulation, and backpropagation. During the selection phase, starting from the root node, the best child node is added to the path based on its UCT value, using an exploration weight e of 2.5. In the expansion phase, the current node is expanded according to the expansion width, generating new heuristics that are input to the base model for improvement suggestions. Then, the optimizer summarizes these suggestions.

As shown in Tables~\ref{tab: tsp_prompt} and \ref{tab: FSSP_prompt}, the state transition prompt includes the heuristic of the expanded node, the trajectory of the heuristics, and the improvement suggestions. These are input to the optimizer to generate new heuristic nodes. If a new node is not a terminal node, it is evaluated and added as a child of the expanded node. Each expansion generates new heuristics according to the width of the expansion. During the simulation phase, the last node in the path is recursively expanded, and the node with the highest reward is selected to be added to the path. The simulation ends when the last node meets the terminal condition or an early stopping condition. During backpropagation, the sum of rewards from the node to the leaf/terminal node is appended to the accumulated reward list of the node, from the leaf node back to the root node. The average of these accumulated rewards is then used as the Q-value of the node. To improve computational efficiency and avoid unnecessary exploration of unpromising paths, PoH employs an early stopping mechanism after a depth exceeding 2. Specifically, early stop occurs if a state’s reward falls below a minimum threshold or exceeds a maximum threshold. The minimum threshold is defined as the average of the rewards obtained by the parent node and the root node, while the maximum threshold is the maximum reward observed among all the nodes currently explored. This strategy encourages the discovery of shorter paths within the heuristic search space, thus enhancing overall efficiency.

Each MCTS iteration generates a path from the root node to a leaf node, resulting in dozens of nodes after the search process. Finally, the path with the highest average reward is selected, and the heuristic with the highest reward within that path is chosen as the final output. This strategy is motivated by the fact that the path with the highest average reward represents the best overall search trajectory, and the best heuristic may not always be the last node on that path due to the depth limit causing premature termination.
\subsection{Baseline details}
In our experiments, we detail the specifics of various baseline methods to facilitate a more accurate comparison and evaluation of their performance in heuristic optimization. The following provides detailed descriptions of the three main baseline methods:
\noindent \textbf{Monte Carlo~(MC).}
The MC method is a random sampling-based optimization strategy that performs multiple single-step samplings and selects the best sampled heuristic. It employs the same heuristic sampling method as PoH~(Employing MCTS) but limits the search depth to one step. Although MC is advantageous due to its simple implementation and broad applicability, its convergence speed can be slow, especially with a limited number of samples. To ensure result reliability in our experiments, we sampled 72 new heuristics for each task.

\noindent \textbf{Beam Search.}
Beam search is a tree-structured search algorithm that explores potential heuristics by expanding nodes layer by layer. In our experiments, the beam search uses the same expansion function as PoH. With a beam width of 3, each node~(excluding the root) expands into 3 new nodes. This results in 9 nodes at each level of the search tree, of which the best 3 are retained for subsequent expansion. The root node expands to 9 new nodes. Given a search depth of 8, a total of 72 nodes are generated, representing new heuristic algorithms. By constraining the beam width and search depth, the beam search efficiently explores the search space within limited computational resources.

\noindent \textbf{Greedy Search.}
Greedy search is an optimization method derived from beam search with a beam width of 1, effectively transforming it into a depth-first greedy search. At each step, greedy search selects the currently optimal node for expansion, rapidly converging towards a local optimum. We conducted experiments with the same search depth of 8 but explored different expansion widths. For instance, Greedy with an expansion width of 3 generates 34 heuristics. While greedy search offers high computational efficiency, it is susceptible to becoming trapped in local optima, particularly when the heuristic function is not sufficiently accurate.
\subsection{Explanations on reward function setup}
Regarding the reward function setup, we use current baseline algorithms (e.g., Concorde or LKH for the
TSP) on a pre-generated training set to get the shortest distance (optimal value). During training, we replace GLS’s update
distance matrix function with code generated by the LLM, getting a distance value (evaluation value). To keep the reward
between 0 and 1, and ensure higher rewards are better, we apply the formula: $R = 1 - \frac{O_{st} - O_{best}}{O_{best}}$ Thus, the optimal solution
is the optimal value obtained by the pre-selected baseline algorithm on the training set, this reward function setting method is applicable to most problems, which also does not affect the limitations
of our method. 
\subsection{More details on how the actions are generated and motivation}
To better illustrate how LLMs can generate heuristic for solving combinatorial optimization problems and how to evaluate the generated heuristic, we provide the following details.

Taking the Guided Local Search (GLS) algorithm for solving the Traveling Salesman Problem (TSP) as an example, the GLS algorithm is prone to falling into local optima during the local search phase. To escape this dilemma, it updates the distance matrix function to guide the search direction. By designing different distance matrix update functions, GLS can be guided to enhance exploration, adapt to dynamic changes, improve algorithm efficiency, and balance exploration and exploitation. The essence of heuristic lies in guiding the search process based on experience or specific rules to increase the likelihood of finding satisfactory solutions, rather than guaranteeing global optimality. Therefore, in the context of GLS solving TSP instances, designing the distance matrix update function is also a form of heuristic design.

We describe the specific process of the PoH method as follows: PoH first use LLM to generate (or initialize) a heuristic based on the prompt. Subsequently, the code generated by the LLM is processed to extract the distance matrix update function (achieved using Python's re.findall function), which is then dynamically loaded as a module (using the importlib.import\_module function). Next, the generated distance matrix update function is combined with the GLS algorithm to solve TSP instances, and the performance of the generated heuristic is evaluated (i.e., the reward value of the current heuristic). In the action generation phase, the LLM analyzes the current and past heuristic (states), including their code, descriptions, and evaluation results, to provide improvement suggestions for the current heuristic, thereby generating better heuristic in subsequent iterations. The motivation behind this mechanism is to encourage the LLM to engage in self-reflection, leveraging its strong language and code generation capabilities to continuously optimize heuristic.
\subsection{Distance Matrix Update in GLS}
In TSP, we replace GLS's distance matrix update function with code generated by the LLM. More specifically, taking the Guided Local Search (GLS) algorithm for solving the Traveling Salesman Problem (TSP) as an example, the GLS algorithm is prone to falling into local optima during the local search phase. To escape this dilemma, it updates the distance matrix function to guide the search direction. By designing different distance matrix update functions, GLS can be guided to enhance exploration, adapt to dynamic changes, improve algorithm efficiency, and balance exploration and exploitation. The essence of heuristic lies in guiding the search process based on experience or specific rules to increase the likelihood of finding satisfactory solutions, rather than guaranteeing global optimality. Therefore, in the context of GLS solving TSP instances, designing the distance matrix update function is also a form of heuristic design.
\subsection{Guided Local Search}
Guided Local Search~(GLS) is a widely adopted strategy to guide local search away from local optima in combinatorial optimization problems. When a typical local search becomes trapped in a local optimum, GLS modifies the objective function to direct the search toward more promising regions. Our objective is to leverage PoH to discover effective heuristics to enhance GLS. In our experimental setup, we employ a variant of the classic GLS algorithm that incorporates a perturbation phase [1], where edges with higher heuristic values are preferentially penalized. During training, we used TSP200 with 800 GLS iterations to evaluate each heuristic.
\subsection{Prompt Engineering}

This section details the prompt formats used in Planning of Heuristics~(PoH). As shown in Tables~\ref{tab: tsp_prompt} and \ref{tab: FSSP_prompt}, the “example\_string” represents the code of each heuristic example. The “improvement suggestion” includes several heuristic code examples and guides the optimizer model to generate improvement suggestions. The “state transition” prompts the optimizer model to perform state transitions~(i.e., generate new heuristics), including information on the heuristic code examples and the sequence of heuristics on the selected path, known as the “trajectory heuristics.”

\begin{table*}[ht]
\centering
\caption{The prompt of PoH for TSP task}
\label{tab: tsp_prompt} 
\centering
\resizebox{0.9\textwidth}{!}{%
\begin{tabular}{@{}lp{10cm}}
\toprule
Prompt for Initialization    & You are an expert in the domain of optimization heuristics . Your task is to design heuristics that can effectively solve optimization problems.
\textbf{Firstly},describe your new heuristic in one
sentence. The description must start with \textless{}start\textgreater{} and end with \textless{}end\textgreater{}. 

\textbf{Next}, implement it in Python as a function named 'heuristics'.The 'heuristics' function takes as input a distance matrix, and returns prior indicators of how bad it is to include each edge in a solution. The return is of the same shape as the input.\par
All inputs and outputs are Numpy arrays.
Do not give additional explanations.  \\ 
\midrule
example\_string    & \textless\{index\}\textgreater \par 
The heuristics code is:
\{algorithm\} \par
The reward of the heuristic is: \{reward\}. \\ 
\midrule
improvement suggestions       &  You are an expert in the domain of optimization heuristics . Your task is to give hints to design better heuristics.\par
My current heuristics is:
\{cur\_algorithm\} \par
This heuristics is not good enough:
\{example\_string\} \par
For each example, the reward higher the heuristics better, identify the common idea in the provided heuristics. Finally, based on all these reasons, summarize and list all the aspects that can improve the heuristics.          \\
\midrule
state\_transit         &    You are an expert in the domain of optimization heuristics . Your task is to give hints to design better heuristics.\par
Here are some algorithm and the corresponding code and the reward is:
\{example\_string\} \par
Based on these examples, the improvement of suggestion with this heuristic and the reasons are:
\{gradient\} \par
There are a list of former algorithms including the current heuristics, and each heuristic is modified from its former heuristics,the reward higher the heuristic better:
\{trajectory\_prompts\} \par
Based on the above information, Please help me design a new heuristic that is different from the given ones
but can be motivated by them.
\textbf{Firstly}, identify the common idea in the provided heuristics.\par
\textbf{Secondly}, based on the backbone idea describe your new heuristic in one
sentence. The description must start with \textless{}start\textgreater{} and end with \textless{}end\textgreater{}. \par
\textbf{Thirdly}, implement it in Python as a function named 'heuristics'.The 'heuristics' function takes as input a distance matrix, and returns prior indicators of how bad it is to include each edge in a solution. The return is of the same shape as the input.\par
All inputs and outputs are Numpy arrays. Do not give additional explanations. \\  \bottomrule
\end{tabular}
}
\end{table*}
\subsection{Generated heuristic}
\begin{figure}[t]
\centering
\includegraphics[width=0.8\columnwidth]{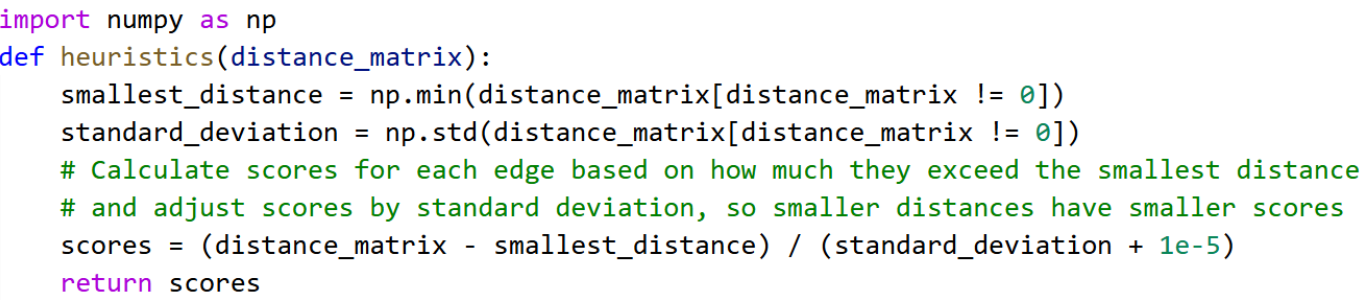} 
\caption{Heuristic generated by PoH with gpt-4.0 on TSP task.}
\label{generated tsp}
\end{figure}

Figure~\ref{generated tsp} shows the heuristic optimized by PoH for updating the distance matrix for the TSP. This heuristic evaluates edges by combining their relative distance from the shortest edge with an adaptive weighting. This weighting prioritizes shorter edges while maintaining diversity based on the standard deviation of all distances.
\subsection{More results}
As shown in Table~\ref{table:TSPLibresults}, for a fair comparison, we used the same TSPLIB test set (containing instances with fewer than 200 nodes) as EoH. The results are presented in Table~\ref{table:TSPLibresults}. In the main text, we also present results for other large-scale instances of TSPLIB, comparing PoH with existing LLM-based heuristic optimization methods.
\begin{table*}[ht!]
\centering
\caption{Results on TSPLib instances. The gap (\%) to the best-known solution from TSPLib.}~\label{table:TSPLibresults}
\large
\renewcommand\arraystretch{0.95}
\resizebox{0.9\textwidth}{!}{%
\begin{tabular}{cccccccccccc}
\toprule
\multirow{2}{*}{Instance} & \multicolumn{3}{c}{Other Algorithms} & \multicolumn{6}{c}{GLS Algorithms} & \multirow{2}{*}{EoH}&\multirow{2}{*}{PoH} \\
\multicolumn{1}{c}{}  & AM& POMO & LEHD& GNNGLS & NeuralGLS & LS    & GLS   & EBGLS & KGLS  &  
\\
\midrule
eil51    & 1.63   & 0.83  & 1.64  & 0.00 & 0.00 & 2.85 & 0.67 & 0.67 & 0.67 & 0.67 & 0.67\\
berlin52 & 4.17   & 0.04  & 0.03  & 0.14 & 0.00 & 3.89 & 0.03 & 0.03 & 0.03 & 0.03& 0.03 \\
st70     & 1.74   & 0.31  & 0.33  & 0.76 & 0.00 & 2.64 & 0.31 & 0.31 & 0.31 & 0.31& 0.31 \\
eil76    & 1.99   & 1.18  & 2.54  & 0.16 & 0.00 & 3.93 & 1.37 & 1.18 & 1.18 & 1.48 & 1.18 \\
pr76     & 0.82   & 0.00  & 0.22  & 0.04 & 0.82 & 6.71 & 0.00 & 0.00 & 0.00 & 0.00 & \textbf{0.00}\\
rat99    & 2.65   & 2.39  & 1.10  & 0.55 & 0.72 & 6.58 & 1.55 & 0.74 & 0.68 & 0.68 & 0.68\\
kroA100  & 4.02   & 0.41  & 0.12  & 0.73 & 0.03 & 3.00 & 0.02 & 0.02 & 0.06 & 0.02 &0.02\\
kroB100  & 5.14   & 0.32  & 0.26  & 0.15 & 0.88 & 0.58 & 0.23 & 0.00 & 0.25 & 0.00 &\textbf{0.00}\\
kroC100  & 0.97   & 0.18  & 0.32  & 1.57 & 1.77 & 4.70 & 0.50 & 0.01 & 0.01 & 0.01 &\textbf{0.01}\\
kroD100  & 2.72   & 0.84  & 0.38  & 0.57 & 0.00 & 5.67 & 0.00 & 0.20 & 0.00 & 0.00 &\textbf{0.00}\\
kroE100  & 1.47   & 0.45  & 0.43  & 1.22 & 1.05 & 4.64 & 0.49 & 0.00 & 0.07 & 0.14 &0.05\\
rd100    & 3.41   & 0.01  & 0.01  & 0.46 & 0.00 & 1.27 & 0.01 & 0.01 & 0.02 & 0.01 &\textbf{0.01}\\
eil101   & 2.99   & 1.84  & 2.31  & 0.20 & 0.36 & 8.82 & 3.28 & 1.91 & 2.07 & 2.27 & 1.78\\
lin105   & 1.74   & 0.52  & 0.34  & 0.61 & 0.65 & 1.87 & 0.03 & 0.03 & 0.03 & 0.03 &\textbf{0.03}\\
pr107    & 3.93   & 0.52  & 11.24 & 0.44 & 0.81 & 0.72 & 0.40 & 0.00 & 0.00 & 0.00 &\textbf{0.00}\\
pr124    & 3.68   & 0.60  & 1.11  & 0.76 & 0.08 & 2.44 & 0.60 & 0.60 & 0.08 & 0.00 &\textbf{0.00}\\
bier127  & 5.91   & 13.72 & 4.76  & 1.95 & 2.73 & 1.79 & 0.59 & 0.29 & 0.42 & 0.42 &\textbf{0.01}\\
ch130    & 3.18   & 0.16  & 0.55  & 3.52 & 1.19 & 7.61 & 1.09 & 0.46 & 0.01 & 0.01 &\textbf{0.01}\\
pr136    & 5.06   & 0.93  & 0.45  & 3.39 & 2.32 & 6.30 & 2.01 & 0.28 & 0.24 & 0.00&\textbf{0.00}\\
pr144    & 7.64   & 0.53  & 0.19  & 3.58 & 0.74 & 4.19 & 0.09 & 0.00 & 0.00 & 0.00&\textbf{0.00} \\
ch150    & 4.58   & 0.53  & 0.52  & 2.11 & 2.49 & 1.35 & 0.68 & 0.37 & 0.04 & 0.24 &0.31\\
kroA150  & 3.78   & 0.70  & 1.40  & 2.98 & 0.77 & 5.05 & 1.75 & 0.26 & 0.17 & 0.00 &\textbf{0.00}\\
kroB150  & 2.44   & 1.17  & 0.76  & 3.26 & 3.11 & 5.55 & 1.01 & 0.00 & 0.08 & 0.00 &\textbf{0.00}\\
pr152    & 7.49   & 1.05  & 12.14 & 3.12 & 0.00 & 2.75 & 0.19 & 0.19 & 0.19 & 0.19 &0.19\\
u159     & 7.55   & 0.95  & 1.13  & 1.02 & 0.90 & 5.63 & 0.74 & 0.78 & 0.96 & 0.00 &\textbf{0.00}\\
rat195   & 6.89   & 8.15  & 1.42  & 1.67 & 0.48 & 2.14 & 0.61 & 0.61 & 0.97 & 0.82&0.62 \\
d198     & 373.02 & 17.29 & 9.23  & 4.77 & 1.28 & 7.96 & 2.08 & 1.87 & 0.31 & 0.59   &0.32\\
kroA200  & 7.11   & 1.58  & 0.64  & 2.03 & 0.86 & 0.91 & 0.75 & 0.18 & 0.71 & 0.15&\textbf{0.05} \\
kroB200  & 8.54   & 1.44  & 0.16  & 2.59 & 3.74 & 4.71 & 1.43 & 1.27 & 0.89 & 0.20  &\textbf{0.01} \\
\midrule
Average  & 16.77  & 2.02  & 1.92  & 1.53 & 0.96 & 4.01 & 0.78 & 0.42 & 0.36 & 0.30 & \textbf{0.22}\\
\bottomrule
\end{tabular}%
}
\end{table*}
\subsection{Guided Local Search}

We employ the same GLS framework previously used for the Traveling Salesman Problem~(TSP) and have selected two common local search operators: Swap and Relocate. We then apply the PoH methodology to design a specialized heuristic strategy for two key tasks within the GLS framework: (1) dynamically updating the execution time matrix and (2) identifying the set of jobs to be perturbed.
\subsection{Prompt Engineering}
As shown in table~\ref{tab: FSSP_prompt}, the prompting engineering framework for the FSSP is almost identical to that of the TSP, the key difference is the objective: designing a heuristic specifically tailored for the FSSP.
\subsection{Generated heuristic}
\begin{figure}[!t]
\centering
\includegraphics[width=0.8\columnwidth]{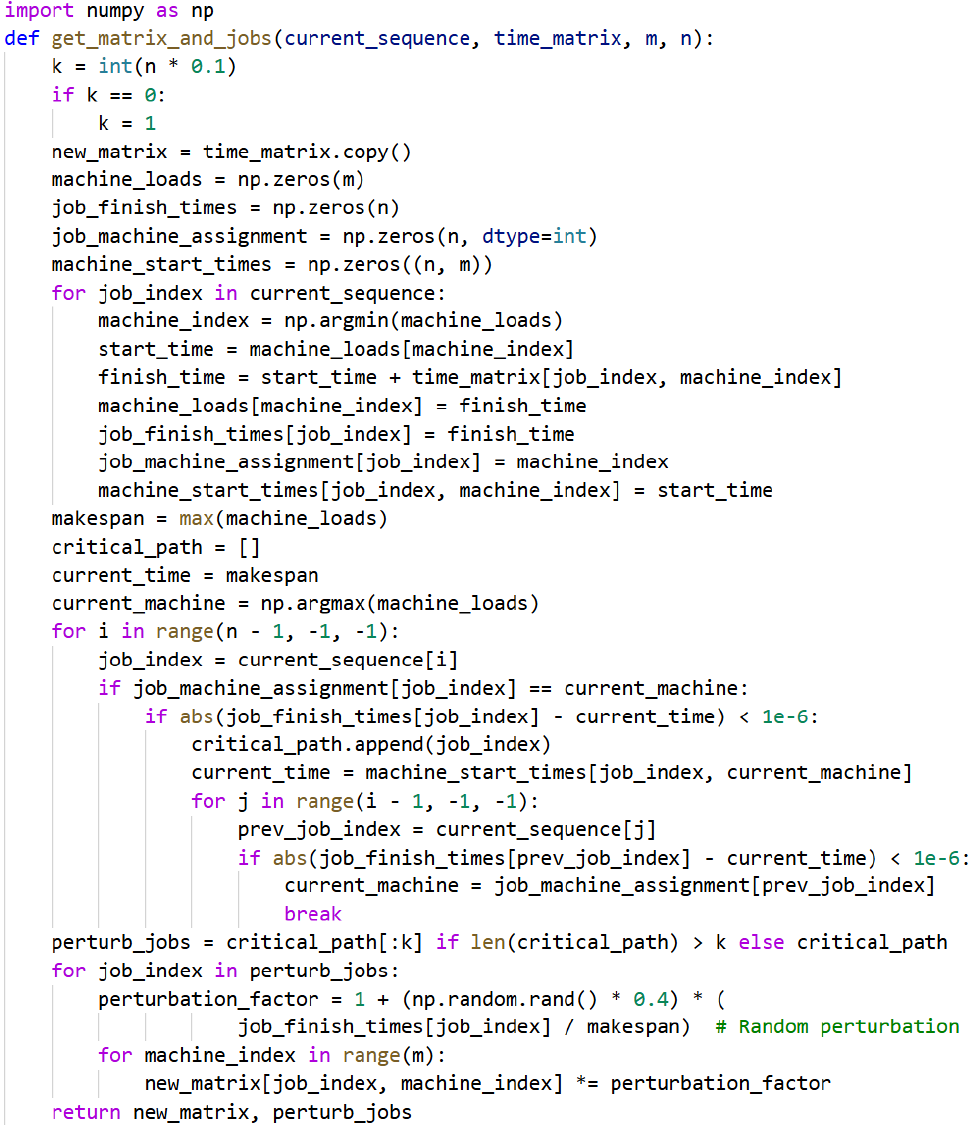} 
\caption{Heuristic generated by PoH with gpt-4.0 on FSSP task.}
\label{jssp}
\end{figure}

Figure~\ref{jssp} shows the PoH-optimized heuristic for updating both the execution time matrix and determining the perturbed jobs for the FSSP. This heuristic perturbs the execution times of jobs on the critical path by a randomly sampled factor within a specified range, proportional to their contribution to the makespan, and selects these jobs for perturbation.
\begin{table}[htb]
\centering
\caption{The prompt of PoH for FSSP task}
\label{tab: FSSP_prompt} 
\centering
\resizebox{0.78\textwidth}{!}{%
\begin{tabular}{@{}lp{10cm}}
\toprule
Prompt for Initialization    & You are an expert in the domain of optimization heuristics. I have n jobs and m machines.Your task is to design heuristics that to update the execution time matrix and select the top jobs to perturb to avoid being trapped in the local optimum scheduling with the final goal of finding scheduling with minimized makespan.

\textbf{Firstly},describe your new heuristic in one sentence. The description must start with \textless{}start\textgreater{} and end with \textless{}end\textgreater{}.

\textbf{Secondly}, implement it in Python as a function named ‘get\_matrix\_and\_jobs’. This function should accept four inputs: ”current\_sequence”,”time\_matrix”,”m”,”n”.The function should return two output:”new\_matrix”,’perturb\_jobs’. The variable 'current\_sequence' represents the current sequence of jobs. The variables 'm' and 'n' denote the number of machines and number of jobs, respectively. \
The variable 'time\_matrix' is a matrix of size n*m that contains the execution time of each job on each machine. The output 'new\_matrix' is the updated time matrix, and 'perturb\_jobs' includes the top jobs to be perturbed.

The matrix and job list are Numpy arrays. Do not give additional explanations. 
 \\ 
\midrule
example\_string    & \textless\{index\}\textgreater \par 
The heuristics code is:
\{algorithm\} \par
The heuristics's reward is: \{reward\}. \\ 
\midrule
improvement suggestions       &  You are an expert in the domain of optimization heuristics . Your task is to give hints to design better heuristics.\par
My current heuristics is:
\{cur\_algorithm\} \par
This heuristics is not good enough:
\{example\_string\} \par
For each example, the reward higher the heuristics better, identify the common idea in the provided heuristics. At last, based on all these reasons, summarize and list all the aspects that can improve the the heuristics.          \\
\midrule
state\_transit         &    You are an expert in the domain of optimization heuristics . Your task is to give hints to design better heuristics.\par
Here are some algorithm and the corresponding code and the reward is:
\{example\_string\} \par
Based on these examples, the improvement of suggestion with this heuristic and the reasons are:
\{gradient\} \par
There are a list of former algorithms including the current heuristics, and each heuristic is modified from its former heuristics,the reward higher the heuristic better:
\{trajectory\_prompts\} \par
Based on the above information, Please help me design a new heuristic that is different from the given ones
but can be motivated by them.
\textbf{Firstly}, identify the common idea in the provided heuristics.\par
\textbf{Secondly}, based on the backbone idea describe your new heuristic in one
sentence. The description must start with \textless{}start\textgreater{} and end with \textless{}end\textgreater{}. \par
\textbf{Thirdly}, implement it in Python as a function named ‘get\_matrix\_and\_jobs’. This function should accept four inputs: ”current\_sequence”,”time\_matrix”,”m”,”n”.The function should return two output:”new\_matrix”,’perturb\_jobs’. The variable 'current\_sequence' represents the current sequence of jobs. The variables 'm' and 'n' denote the number of machines and number of jobs, respectively. \
The variable 'time\_matrix' is a matrix of size n*m that contains the execution time of each job on each machine. The output 'new\_matrix' is the updated time matrix, and 'perturb\_jobs' includes the top jobs to be perturbed.

The matrix and job list are Numpy arrays. Do not give additional explanations.  \\  \bottomrule
\end{tabular}
}
\end{table}
\subsection{More results}
As shown in Table~\ref{table:taillard_Appendix}, using the complete set of Taillard instance (with the number of jobs ranging from 20 to 200 and the number of machines ranging from 5 to 20), which include 10 instances at each of 11 different scales, PoH outperforms EoH in most cases.
\begin{table}[ht]
\centering
\caption{Results on Taillard instance sets. The value is the average gap to the best-known solutions on 10 instances in each set. The best results are in bold. }~\label{table:taillard_Appendix}
\large
\renewcommand\arraystretch{0.95}
\resizebox{0.7\textwidth}{!}{
\begin{tabular}{lcccccccccc}
\toprule
  Test Set& GUPTA & CDS   & NEH  & NEHFF & PFSPNet & LS   & ILS1 & ILS2  & EoH &PoH   \\
\midrule
20\_5   & 12.89 & 9.03  & 3.24 & 2.30  & 2.30 & 1.91 & 0.42 & 0.18  & \textbf{0.09}   & 0.10\\
20\_10  & 23.42 & 12.87 & 4.05 & 4.15  & 4.04 & 2.77 & 0.33 & 0.25 & 0.30& \textbf{0.19}  \\
20\_20  & 21.79 & 10.35 & 3.06 & 2.72  & 2.96 & 2.60 & 0.29 & 0.25  & \textbf{0.10} & 0.22 \\
\midrule
50\_5   & 12.23 & 6.98  & 0.57 & 0.40  & 0.51 & 0.32 & 0.15 & 0.32  & \textbf{0.02} & \textbf{0.02} \\
50\_10  & 20.11 & 12.72 & 3.47 & 3.62  & 3.48 & 3.33 & 1.47 & 0.29  & \textbf{0.19}  & \textbf{0.19}\\
50\_20  & 22.78 & 15.03 & 5.48 & 5.10  & 5.05 & 4.67 & 2.13 & \textbf{0.34} & 0.60   & 0.49\\
\midrule
100\_5  & 5.98  & 5.10  & 0.39 & 0.31  & 0.31 & 0.28 & 0.20 & 0.38  & \textbf{-0.04}& 0.06 \\
100\_10 & 15.03 & 9.36  & 2.07 & 1.88  & 1.72 & 1.38 & 0.77 & 0.34  & 0.14 & \textbf{0.11} \\
100\_20 & 21.00 & 13.55 & 3.58 & 3.73  & 3.56 & 3.51 & 2.27 & 0.43 & 0.41& \textbf{0.38}  \\
200\_10 & 11.59 & 7.22  & 0.98 & 0.70  & 0.82 & 0.87 & 0.74 & 0.54  & \textbf{0.12} &\textbf{0.12}  \\
\midrule
200\_20 & 18.09 & 11.89 & 2.90 & 2.52  & 2.49 & 2.53 & 2.26 & 0.59 & 0.61  & \textbf{0.53}\\
\midrule
Average & 16.81 & 10.37 & 2.71 & 2.49  & 2.48 & 2.20 & 1.00 & 0.36  & 0.23 & \textbf{0.22}\\
\bottomrule
\end{tabular}%
}
\end{table}


\end{document}